\pdfoutput=1

\documentclass[11pt]{article}

\usepackage[]{ACL2023}

\usepackage{times}
\usepackage{latexsym}

\usepackage[T1]{fontenc}

\usepackage[utf8]{inputenc}

\usepackage{microtype}

\usepackage{inconsolata}

\usepackage{amsmath, amssymb, amscd, amsthm, amsfonts}
\usepackage{mathtools,enumerate,mathrsfs,graphicx}
\usepackage{subfigure}
\usepackage{multirow}
\usepackage{hyperref}
\usepackage{algorithm}
\usepackage{algorithmicx}
\usepackage[inline]{enumitem}
\usepackage{algpseudocode}
\usepackage{stfloats}
\usepackage{tcolorbox}
\usepackage{booktabs}
\usepackage{amsmath}
\usepackage{float}

\usepackage{microtype}

\usepackage{sistyle}
\SIthousandsep{,}

\title{One-stop Training of Multiple Capacity Models }

\author{
Lan Jiang$^{1}$\thanks{\ \ Equal contribution}, Haoyang Huang$^{2}$\footnotemark[1], Dongdong Zhang$^{2}$, Rui Jiang$^{1}$, \textbf{Furu Wei$^{2}$\thanks{\ \ Corresponding Author}}  \\
$^1$MOE Key Laboratory of Bioinformatics, Center for Synthetic and Systems Biology, \\
Department of Automation, BNRist, Tsinghua University, China \\
$^2$Microsoft Research Asia, China \\
}

\begin{document}
\maketitle

\begin{abstract}

Training models with varying capacities can be advantageous for deploying them in different scenarios. 
While high-capacity models offer better performance, low-capacity models require fewer computing resources for training and inference. 
In this work, we propose a novel one-stop training framework to jointly train high-capacity and low-capactiy models. 
This framework consists of two composite model architectures and a joint training algorithm called Two-Stage Joint-Training (TSJT).
Unlike knowledge distillation, where multiple capacity models are trained from scratch separately, our approach integrates supervisions from different capacity models simultaneously, leading to faster and more efficient convergence.
Extensive experiments on the multilingual machine translation benchmark WMT10 show that our method outperforms low-capacity baseline models and achieves comparable or better performance on high-capacity models. 
Notably, the analysis demonstrates that our method significantly influences the initial training process, leading to more efficient convergence and superior solutions.

\end{abstract}
\section{Introduction}
Scaling up model capacities has become a promising strategy for improving performance on various natural language processing benchmarks.
However, increasing the number of parameters in Large Language Models (LLMs)~\cite{chowdhery2022palm,brown2020language,openai2023gpt4} and Sparse Mixture-of-Experts (SMoE) models~\cite{fedus2022switch,lepikhin2020gshard,zuo2022taming,dai-etal-2022-stablemoe} can result in extremely high computational costs and difficulties in fine-tuning them. 
As a result, researchers have been exploring low-capacity models~\cite{park2021learning,jiang-etal-2022-rose,xu_survey_2022} as an alternative.
Additionally, recent studies have shown that low-capacity models~\cite{mirzadeh2020improved,xu2023small} can also be collaboratively fine-tuning high-capacity models and valuable plug-ins for Large Language Models. 
Therefore, there is still a significant need for jointly training multi-capacity models in real-world applications.

The encoder-decoder framework with varying capacities has been extensively employed in numerous NLP generation tasks, particularly for multilingual machine translation ~\cite{vaswani2017attention,lewis2019bart,raffel2020exploring,xue2020mt5}. 
However, there is a lack of research on whether high-capacity and low-capacity models can promote each other in this task. 
The traditional approach of involving multiple capacity models is knowledge distillation technique (KD).
It has been developed to distill knowledge from high-capacity models and improve low-capacity models, resulting in notable success~\cite{tang2019distilling,NEURIPS2019_2c601ad9,sun-etal-2020-knowledge-distillation,rao2022parameter}. 
Nevertheless, this method still has two main drawbacks.
Firstly, the serial training pipeline requires high-capacity models to be prepared before low-capacity models, which increases the overall time cost. 
Secondly, the knowledge distillation process is unidirectional, where the low-capacity models receive useful information from the high-capacity models, but not vice versa.

In this work, we present a novel one-stop training framework of multiple capacity models to address above challenges.
The intuition behind our method is straightforward: to leverage the strengths of models wtih different capacities to facilitate each other, thereby enabling them to find optimal solutions collaboratively, rather than relying solely on individual learning. 
Specifically, we propose a novel joint training algorithm, called Two-Stage Joint-Training (TSJT), which is designed to jointly train high-capacity and low-capacity model leading more efficient convergence.
To further evaluate the effectiveness of TSJT, we introduce two composite model variants, namely shared and indep architecture. 
These two architectures take into account the variety of model capabilities and the extent of shared data.
In addition, TSJT divides the training process into two stages. 
In the first stage, the submodels work collaboratively to integrate supervisions from each other, ultimately reaching their optimal checkpoint. 
In the second stage, TSJT empowers submodels to optimize independently and seek their individual optimal solutions.

We conduct extensive experiments on the multilignual machine translation benchmark WMT10 to evaluate the effectiveness of our one-stop training schema.
The resutls show that our method exhibits superior performances on relatively low-capacity models, and achieves comparable or even better performance on high-capacity models.
Furthermore, we delve into a detailed analysis of our approach by closely examining the optimization trajectory and loss visualizations. 
The results demonstrate that TSJT has a significant impact on the initial training process, leading models to converge faster and reduce the overall training time cost. 
Subsequently, TSJT empowers models to identify their best possible solutions, while ensuring stability and keeping the loss minimal.

\section{One-stop Training Framework}

In this section, we will introduce one-stop training framework in detail.
Our method includes three critical components: the multiple capacity model architecture with shared or independent layers (Section~\ref{sec:method-arch}), the two-stage joint training algorithm (Section~\ref{sec:method-asy}), and the training objective (Section~\ref{sec:method-obj}) which optimizes all capacity models simultaneously during training.
The multi-capacity model architectures enables submodels with adaptable depth and width.
Once the one-stop training finished, various capacity models can be extracted and utilized.
The two-stage joint training algorithm makes the best of joint training supervision.
It makes models with varying capacities to achieve faster and better convergence at the initial training process, as well as allows them to explore their own optimal solutions. 
The training objective determine specific optimization target in different training stage, with or without additional supervision from other models.

\subsection{Multiple Capacity Models}
\label{sec:method-arch}

\begin{figure*}[htb]
    \centering
    \subfigure[\textbf{Shared} Arch: MoE and dense share layers.]{
		\begin{minipage}[b]{0.482\textwidth}
			\includegraphics[width=\textwidth]{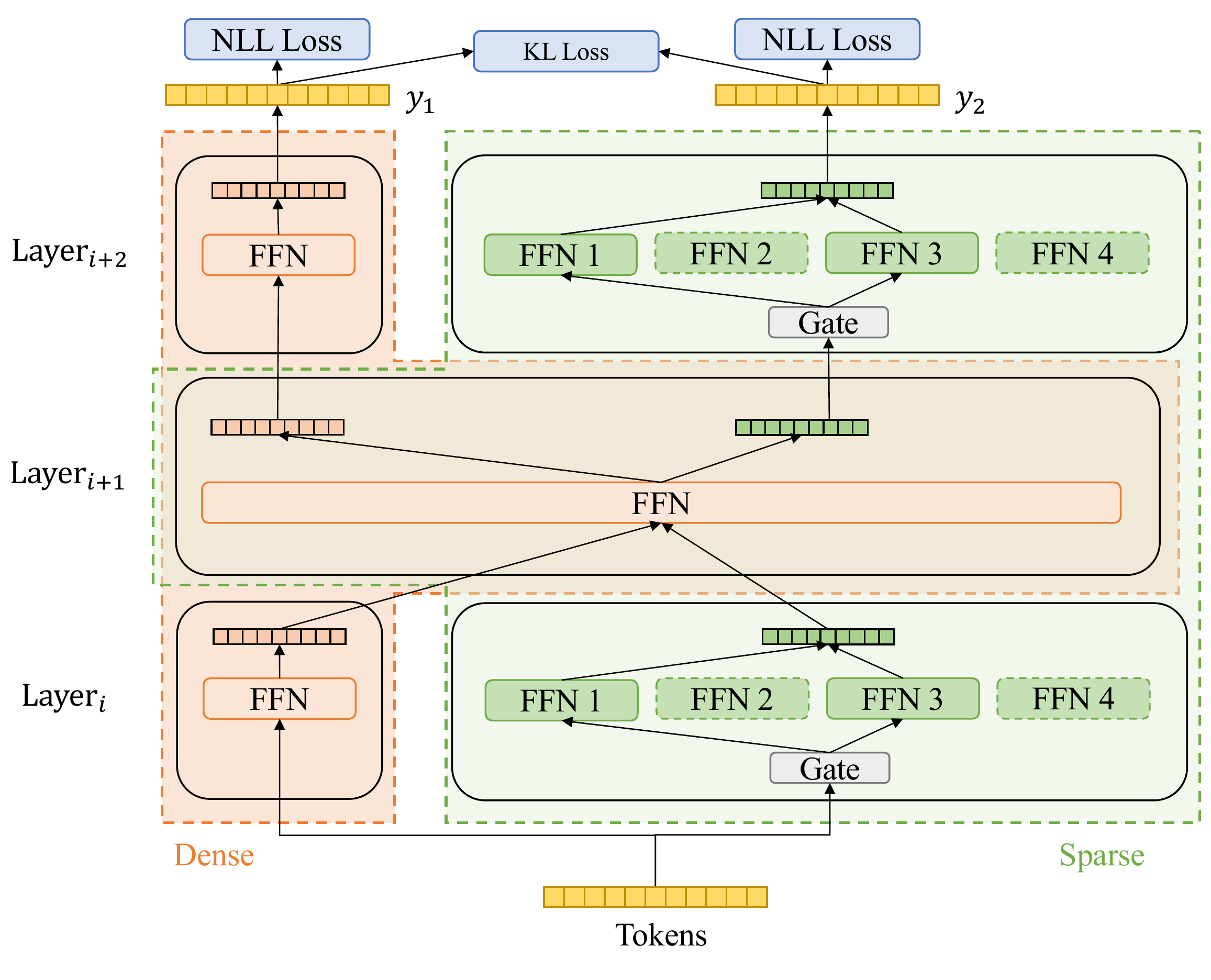}
		\end{minipage}
        \label{fig:arch-share}
	}
	\subfigure[\textbf{Indep} Arch: MoE and device are independent.]{
		\begin{minipage}[b]{0.482\textwidth}
			\includegraphics[width=\textwidth]{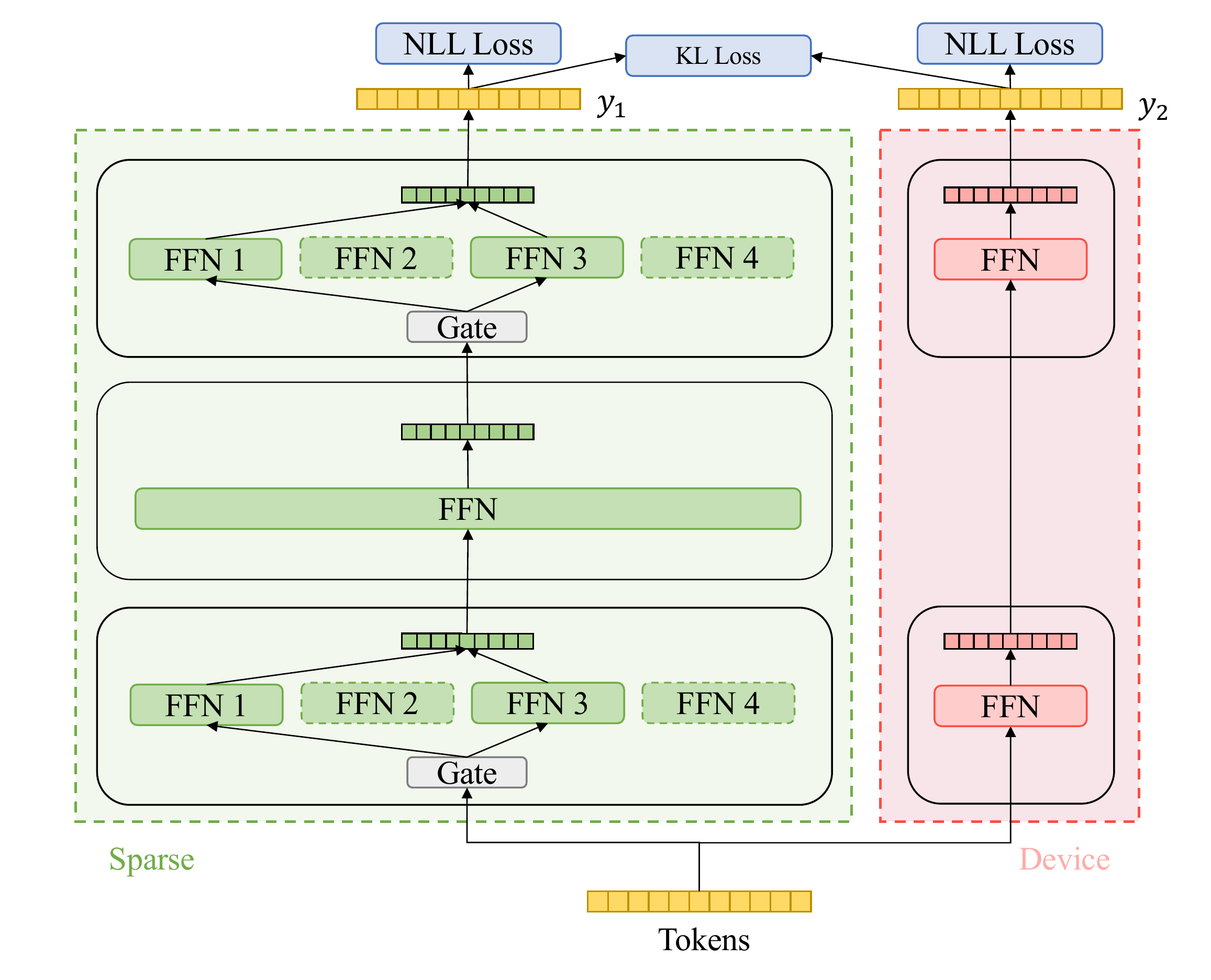}
		\end{minipage}
        }
    \caption{Two model architecture variants in our two-stage joint-training schema.
    \textbf{(a) Shared} architecture variant includes a MoE and a dense model, where they share specific layers. 
    The shared layers are optimized by both of them.
    Thus dense model has limited width (same as MoE), but flexible depth.
    \textbf{(b) Indep}endent architecture variant includes a MoE and a device model, where they are completely independent from each other.
    Device model has flexible width and depth.
    }
    \label{fig:arch}
\end{figure*}

The model architectures in our one-stop training framework is to verify the Multilingual machine translation task, we use the standard encoder-decoder framework for models with different capacities.
Once the training finished, all submodels can be separated from the composite model architecture and utilized.
As shown in Figure \ref{fig:arch}, we propose two variations of model architecture, namely shared and indep (independent) architecture, to cater to the requirements of different capacity models.

\textbf{Shared architecture.} 
The shared architecture provide two submodels with specific shared parameters.
We take MoE and dense model as example in the shared architecture.
The MoE model is consist of moe layer in even-numbered layers and standard transformer layer in odd-numbered layers.
The dense model shares the standard transformer layer with MoE, but possesses unique parameters in its even-numbered layers.
In the forward process, the hidden states of both MoE and dense models go through layers with identical parameters at odd-numbered layers. 
Meanwhile, in the backward pass, the shared layers are jointly optimized by the two submodels.
As a result, submodels within the shared architecture can benefit from the common parameters. However, this sharing necessitates that the two submodels maintain the same width, which constrains the capacity ratio between them.

\textbf{Indep architecture.}
The independent architecture also includes two models with varying capacities. 
The primary distinction between the two architectures is the existence of shared parameters among the submodels within them. 
In the independent architecture, the two submodels are entirely separate from one another.
Consequently, although there may be a loss in sharing information, the independent architecture can offer submodels with a wider range of capacities.
Using the MoE and device model as examples in the independent architecture, the device model typically has half the hidden size and depth of the MoE model. 
Additionally, the device model inserts layers corresponding to the locations of the MoE layers in the sparse model.
It should be noted that within this architectural framework, the high-capacity submodel isn't necessarily required to be a MoE model. 
It could just as well be an arbitrary large model, such as a large-scale pre-trained language model.

Theoretically, these two model architectures allow various backbone submodels with flexible capacity.
Due to the limit of space, we focus on three representative capacity models in this work, which are sparse, dense and device, listed in descending order of their size. 

The sparse model is usually a deep and sparse model.
The capacity of sparse model have direct influence on the following dense and device model.
In this work, we employ a Mixture-of-Experts (MoE) model as sparse model.
Note that arbitrary sparser and deeper models can be adopted as the sparse capacity model here.
The dense capacity model is a relatively compact and small one, similar to a vanilla encoder-decoder model comprising roughly $300$ million parameters.
On the other hand, the device capacity represents the smallest model in our schema, with less than $100$ million parameters. 


\subsection{Two-stage Joint Training Algorithm}
\label{sec:method-asy}

\begin{figure*}[ht]
    \centering
    \includegraphics[width=\linewidth]{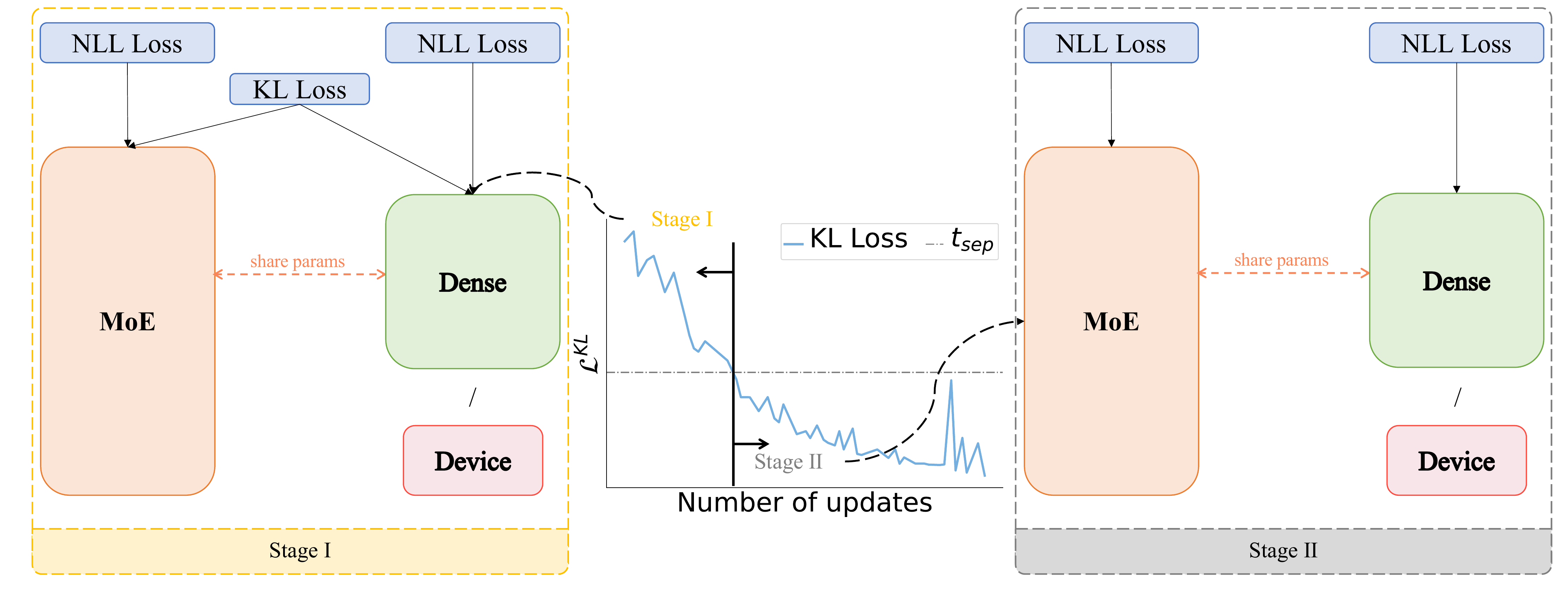}
    \caption{Illustration of \textbf{T}wo-\textbf{S}tage \textbf{J}oint-\textbf{T}raining (\textbf{TSJT}) algorithm for shared and independent model architecture.
    In the first stage two models are trained with additional KL constraint.
    In the second stage two models are trained seperately.
    Note that two models in shared architecture should be updated simultaneously due to the shared parameters.}
    \label{fig:asy}
\end{figure*}

We then propose a two-stage joint training algorithm to train each submodel exhaustively.
The illustration of \textbf{T}wo-\textbf{S}tage \textbf{J}oint-\textbf{T}raing (\textbf{TSJT}) algorithm is shown in Figure~\ref{fig:asy}.

The idea of the TSJT algorithm is similar to the pre-training and fine-tuning of the language model.
The first stage emphasizes global optimization, enforcing consistency constraints between submodels of varying capacities to aid in their quicker and more efficient convergence. 
After reaching the rather optimal region, we transition to the second stage, during which the constraint is removed and the submodels are local fine-tuned to find optimal solution, respectively. 
It is not reasonable to maintain the strong constraint between models of different capacities throughout, as this could hinder their ability to discover optimal solutions.
The timing of the stage transition is determined by the divergence between two submodels.

Specifically, we employ the KL loss $\mathcal{L}^\text{KL}$ of the outputs of two submodels as the quantified divergence, and set a separate threshold $t_{\text{sep}}$.
Once
\begin{eqnarray}
    \label{eq:1}
    \mathcal{L}^\text{KL}\leq t_{\text{sep}},
\end{eqnarray}
the TSJT algorithm completes the first stage and proceeds to the second stage.

During the first stage, the shared and independent architectures must execute the forward process twice to calculate their respective cross-entropy loss and KL loss. 
In each backward process step, we optimize the MoE submodel first, followed by the dense or device submodel. 
In the second stage, the process is nearly identical to the first stage, except that the calculation of KL loss is omitted. 
Notably, during the second stage, the MoE and device submodel from the independent architecture can be trained asynchronously, which is not applicable in the shared architecture.

\subsection{Training Objective}
\label{sec:method-obj}

In this section, we will demonstrate the derivation of our composite learning objective for each submodel within our novel model architectures during the joint training process.
As previously stated, our model architecture comprises two submodels with varying capacities.
Our joint training scheme aims to leverage the strengths of each submodel to complement the other, ultimately enabling them to find optimal solutions collaboratively, rather than relying solely on individual learning.
Specifically, we add a consistency constraints into the original training objective of each submodel in the first stage of TSJT.
Such constrain could employ the knowledge of sparse model to facilitate the learning process of dense and device model, and versa vice.

Using submodels from the shared architecture as an example, the original training objective of models is the cross-entropy loss.
Given a source sequence $\textbf{x}$ of length $S$ and a target sequence $\textbf{y}$ of length $T$, the training objective $\mathcal{L}$ is defined as:

\begin{eqnarray}
\label{eq:2}
    \mathcal{L}=-\frac{1}{T}\sum_{t=1}^{T}\log\mathcal{P}(\mathbf{y}_t\lvert\mathbf{x}).
\end{eqnarray}

While in our schema, the two submodels should also keep consistent with each other during the  training process.
Considering the output of the MoE model $\mathbf{y}$ and the dense model $\mathbf{y}^{'}$, then the Kullback-Leibler (KL) divergence between them can be derived as :

\begin{eqnarray}
\label{eq:3}
    \mathcal{L}^\text{KL} = \mathcal{D}_{KL}(\mathbf{y}\Vert\mathbf{y}^{'}) + \mathcal{D}_{KL}(\mathbf{y}^{'}\Vert\mathbf{y}).
\end{eqnarray}

Finally, the training objective of dense model is defined as:
\begin{eqnarray}
\label{eq:4}
    \mathcal{L}^{'} = \mathcal{L}+\alpha\cdot\mathcal{L}^\text{KL},
\end{eqnarray}
where $\alpha$ is a scaling coefficient hyperparameter to control the effect of $\mathcal{L}^\text{KL}$.

Similarly, the training objective of submodels within the independent architecture can be derived in the same way.
Noted that for the MoE model, we also add the KL divergence into its training objective.
But the $\alpha$ used in Eq. \ref{eq:4} is usually not the same as rather low-capacity models.
While in the second stage, we use the standard training objective in Eq.~\ref{eq:2}.
We will report more details in experimental settings.

\section{Experiments}

\begin{table*}[thbp]
\centering
\resizebox{\linewidth}{!}{
\begin{tabular}{l|ccccc|cccccccccc|c}
\toprule
\textbf{Model} & \textbf{\# Para} & \textbf{\# Emb} & \textbf{\# Enc} & \textbf{\# Dec} & \textbf{\# Exp} & \textbf{Cs}    & \textbf{De}    & \textbf{Et}    & \textbf{Fi}    & \textbf{Fr}    & \textbf{Gu}   & \textbf{Hi}    & \textbf{Lv}    & \textbf{Ro}    & \textbf{Tr}    & \textbf{Avg} \\ \hline 
\hline
\multicolumn{17}{c}{\textbf{X$\rightarrow$En}}\\ \hline
Single MoE    & 845M & \multirow{2}{*}{768} & 12 & 12 & 8 & 31.70 & 38.50 & 24.40 & 25.90 & 33.40 & 20.60 & 15.80 & 26.90 & 33.60 & 19.80 & 27.06 \\
Single dense  & 320M & & 6 & 6 & 1 & 31.10 & 36.60 & 22.80 & 24.80 & 32.50 & 18.60 & 16.20 & 25.80 & 35.70 & 19.20 & 26.33 \\ 
Single device  & 91M & 288 & 3 & 3 & 1 & 25.20	& 29.60	& 15.70	& 18.80	& 26.90	& 11.00	& 10.70	& 18.50	& 27.20	& 12.60	& 19.62 \\ 
\hline
TSJT-shared MoE   & 845M & \multirow{3}{*}{768} & 12 & 12 & 8 & 33.10 & 39.40 & 25.60 & 26.70 & 33.70 & 22.00 & 19.20 & 28.20 & 37.70 & 21.60 & 28.72  \\
TSJT-shared dense  & 320M &  & 6 & 6 & 1 & 31.70 & 37.10 & 23.30 & 24.90 & 32.40 & 18.80 & 16.60 & 25.90 & 35.90 & 20.10 & 26.67 \\ 
TSJT-indep MoE & 845M & & 12 & 12 & 8 & 33.00 & 39.30 & 25.20 & 26.50 & 33.40 & 21.60 & 19.90 & 28.30 & 37.90 & 21.30 & 28.64 \\
TSJT-indep device & 91M & 288 & 3 & 3 & 1 & 25.70 & 29.70 & 16.10 & 19.60 & 27.40 & 10.90 & 10.90 & 19.50 & 28.10 & 12.50 & 20.04 \\ \hline 
\hline 
\multicolumn{17}{c}{\textbf{En$\rightarrow$X}}\\ \hline
Single MoE    & 845M & \multirow{2}{*}{768} & 12 & 12 & 8 & 25.40 & 33.70 & 19.10 & 21.20 & 31.90 & 12.00 & 11.30 & 24.00 & 28.40 & 17.00 & 22.40 \\
Single dense  & 320M & & 6 & 6 & 1 & 25.10 & 31.60 & 16.30 & 19.40 & 30.10 & 8.40 & 11.10 & 21.40 & 26.00 & 13.60 & 20.30 \\ 
Single device  & 91M & 288 & 3 & 3 & 1 & 19.60	& 24.10	& 11.20	& 13.20	& 25.90	& 3.30	& 6.90	& 14.80	& 19.40	& 7.80 & 14.62 \\ \hline
TSJT-shared MoE   & 845M & \multirow{3}{*}{768} & 12 & 12 & 8 & 26.50 & 34.30 & 18.70 & 21.60 & 32.20 & 10.90 & 12.00 & 23.70 & 28.20 & 15.70 & 22.38 \\
TSJT-shared dense  & 320M & & 6 & 6 & 1 & 25.40 & 31.70 & 16.10 & 19.40 & 30.40 & 8.20 & 11.10 & 21.60 & 25.30 & 13.00 & 20.22 \\ 
TSJT-indep MoE & 845M & & 12 & 12 & 1 & 26.20 & 34.10 & 18.60 & 21.70 & 32.30 & 10.80 & 12.10 & 24.10 & 27.80 & 15.60 & 22.33 \\
TSJT-indep device  & 91M & 288 & 3 & 3 & 1 & 19.50 & 24.70 & 11.20 & 13.70 & 25.60 & 3.70 & 6.70 & 14.90 & 19.50 & 7.80 & 14.73 \\ 
\bottomrule
\end{tabular}}
\caption{
Models performance on WMT10 benchmark on X$\rightarrow$En and En$\rightarrow$X seperately.
Values are reported as percentage (\%).
For each model, we report a macro-average.
\# Para is the number of parameters in models.
\# Emb represents the size of embedding used in models.
\# Enc and \# Dec are the number of layer in encoder and decoder respectively.
\# Exp is the number of experts in each MoE layer (if exist).
}
\label{tab:x2x}
\end{table*}

\subsection{Datasets}

We demonstrate the effectiveness of our methodology on multilingual machine translation tasks.
We adopt a prevalent translation benchmark that includes $10$ languages. 

\textbf{WMT10}~\cite{wang-etal-2020-multi} is a benchmark which includes bitext data between English and other $10$ languages: French (Fr), Czesh (Cs), German (De), Finnish (Fi), Latvian (Lv), Estonian (Et), Romanian (Ro), Hindi (Hi), Turkish (Tr) and Gujarati (Gu).
The training set encompasses a total of $32.5$ million sentence pairs. 
To evaluate the models, we merge all parallel corpora into one training set and assess their performance on individual language test sets. 
Finally, we present the case-sensitive, detokenized BLEU scores using the sacreBLEU metric $\footnote{https://github.com/mjpost/sacreBLEU}$.


\subsection{Baselines}

\textbf{Single Mixture of Experts Model.}
We employ the Mixture-of-Experts models as sparse models in our schema, and conduct single model training as the baseline.
The MoE model is consist of an encoder of $12$ layers and an decoder of $12$ layers, incorporating a MoE layer in every alternate layer.
Each MoE layer includes $8$ experts.
The embedding dim is set to $768$.

\textbf{Single dense and device Model.}
We adopt the single dense model as our baselines which contains the same number of parameters with the dense and device models in the shared and indep architecture respectively.
The dense model is composed of a $6$-layer encoder and a $6$-layer decoder, while the device model features a $3$-layer encoder and a $3$-layer decoder.
The width of the dense model aligns with that of the MoE model, whereas for the device model, it is set to 288, which is the smallest valid width. 
Both the single dense and device models are trained starting from scratch.


\subsection{Experimental Settings}
\label{sec:exp-set}

Our implementation of our schema is based on Fairseq library$\footnote{https://github.com/facebookresearch/fairseq}$~\cite{ott2019fairseq}.
Following the Switch Transformers~\cite{JMLR:v23:21-0998}, we adopt top-$1$ gating in our MoE models.
Additionally, we employ a balancing loss alongside the cross-entropy loss to balance the load of various experts in the MoE model.
The balancing loss is multiplied by $0.01$ and added to the total loss.
For training, we use the Adam optimzer with $4000$ warm-up steps, start learning rate of $5e-4$, and inverse square root scheduler proposed in~\citet{JMLR:v21:20-074}.
We accumulate gradients to make an effective batch size of $32,768$ tokens for all models.
For all baselines and our methods, the maximum number of epochs is set to $8$.
For shared or independent structure, $\alpha$ in Eq.~\ref{eq:4} is set to $5$ for MoE, and $10$ for either the dense or device models to maintain an equal magnitude with the balancing loss.

\subsection{Results on WMT10}

We primarily report the results of different models on WMT10 benchmark, wherein the models are evaluated in both translation directions: 'X$\rightarrow$En' and 'En$\rightarrow$X'.
We also report the key model size hyperparameters for comparison.
The overall results are summarized in Table~\ref{tab:x2x}.

We observe that:

(1) By developing a range of models with varying capacities, our method can harness the strengths of each model to deliver superior performance compared to standard individual training. 

(2) Compared to single MoE model, MoE models from both shared structure and independent structure achieve better performance.
Our method significantly enhances performances in the X$\rightarrow$En direction, and achieves competitive results in the En$\rightarrow$X direction.
In particular, MoE from the shared architecture outperforms the single MoE by $1.4\%$ score in the Cs$\rightarrow$En direction.
While the independent one outperforms the single MoE by $4.1\%$ score in the Hi$\rightarrow$En direction.

(3) For the dense model, the shared one exhibits better performance than the single model in the X$\rightarrow$En direction, and also competitive result in the reverse direction.
On X$\rightarrow$En direction, the dense model from the shared architecture improves performance on every language except for Fr.
For instance, both De and Et see a $0.5\%$ score improvement.
While in the reverse direction, the dense model is not so good on several low-resource languages.

(4) The device model is the smallest-capacity model in our setting. 
We can observe that the device model within the independent architecture improves the performance in two directions.
In the X$\rightarrow$En direction, our method achieves an average $0.42\%$ score improvement, and a $0.11\%$ score improvement in the reverse direction.

\section{Analysis}



In this section, we delve deeper into the inner workings of our method. 
Given the substantial cost of conducting experiments with 10 languages, we select a high-medium-low resource combination from the WMT10 benchmark as the basis for our analysis experiment.
We adopt the Fr, Fi and Hi as representations of high, medium and low resource languages, respectively.
On the new subset benchmark, we mainly conduct experiments to compare the following strategies or models:
\begin{itemize}[itemsep = 2pt, topsep = 8pt, leftmargin = 15pt, listparindent=20pt]
    \item \textbf{Single} trains the model without joint training, \textit{i.e.} the vanilla single model.
    \item \textbf{ConstJT-shared/indep} trains models within shared or independent structures with constant constraint along all the training process.
    \item \textbf{TSJT-shared/indep} trains models within using TSJT algorithm.
\end{itemize}

For all the experiments, we mainly follow the setting from Section \ref{sec:exp-set}.
We use dictionary of $3$ languages, and set the maximum of epochs to $3$.
The results are summarized in Table \ref{tab:x2x-3langs}.

\begin{table}
\centering
\resizebox{\linewidth}{!}{
\begin{tabular}{ll|ccc|c}
\toprule
\textbf{Method} & \textbf{Model} & \textbf{Fr} & \textbf{Fi} & \textbf{Hi} & \textbf{Avg} \\ \hline 
\hline
\multicolumn{6}{c}{\textbf{X$\rightarrow$En}} \\ \hline
\multirow{3}{*}{Single} & MoE   &  31.8 & 23.3 & 14.5 & 23.20   \\
 & Dense & 30.1 & 20.8 & 12.6 & 21.17  \\ 
 & Device & 23.9 & 14.4 & 8.5 & 15.60  \\ 
\hline
\multirow{2}{*}{ConstJT-shared} & MoE & 31.3 & 23.2 & 14.6 & 23.03 \\
 & Dense & 30.3 & 21.7 & 12.8 & 21.60  \\ \hline
\multirow{2}{*}{ConstJT-indep} & MoE & 31.1 & 22.9 & 13.4 & 22.47  \\
 & Device & 25.6 & 15.7 & 8.1 & 16.47 \\ \hline
\multirow{2}{*}{TSJT-shared} & MoE & 32.1 & 23.9 & 15.6 & 23.87 \\
& Dense & 30.8 & 22.3 & 14.1 & 22.40  \\ \hline
\multirow{2}{*}{TSJT-indep} & MoE & 32.0 & 23.9 & 15.2 & 23.70   \\
 & Device & 26.4 & 16.6 & 9.5 & 17.50  \\ \hline
 \hline
\multicolumn{6}{c}{\textbf{En$\rightarrow$X}} \\ \hline
\multirow{3}{*}{Single} & MoE & 30.3 & 18.4 & 9.5 & 19.40  \\
 & Dense & 28.7 & 16.1 & 8.5 & 17.77   \\ 
 & Device & 23.2 & 10.0 & 4.5 & 12.57  \\ 
\hline
\multirow{2}{*}{ConstJT-shared} & MoE & 30.1 & 18.2 & 9.4 & 19.23 \\
 & Dense & 28.8 & 16.5 & 8.3 & 17.87 \\ \hline
\multirow{2}{*}{ConstJT-indep} & MoE & 29.9 & 17.6 & 7.7 & 18.40  \\
 & Device & 24.1 & 11.2 & 4.5 & 13.27  \\ \hline
\multirow{2}{*}{TSJT-shared} & MoE & 30.6 & 19.0 & 10.3 & 19.97  \\
& Dense & 29.4 & 17.3 & 8.6 & 18.43  \\ \hline
\multirow{2}{*}{TSJT-indep} & MoE & 30.8 & 19.1 & 9.4 & 19.77  \\
 & Device & 24.6 & 11.5 & 5.0 & 13.70  \\ 
\bottomrule
\end{tabular}}
\caption{The BLEU scores (\%) on 3 languages.}
\label{tab:x2x-3langs}
\end{table}

\paragraph{Results.} We can observe from Table \ref{tab:x2x-3langs} that, TSJT algorithm outperforms the single training and constant training comprehensively.
Compared to the baselines, TSJT-shared structure improves the result of the dense model up to $1.3\%$ score on average in the X$\rightarrow$En direction, and $0.66\%$ score on average in the reverse direction.
MoE models from both shared and independent strcuture get improvement about $0.5\%$ score on average compared to the single MoE model.
Regarding the constant joint training strategy, the results indicate that it does not consistently surpass the baseline; however, it does outperform in several language translation tasks.
However, the ConstJT strategy still somewhat restricts performance improvement in certain situations, such as the MoE model from the ConstJT-shared method. 
Overall, compared to the baseline, the ConstJT strategy demonstrates limited improvement and is not as effective as our TSJT algorithm. 
This underscores the necessity of the two-stage method, as such constraints may limit further progress when models move away from the initial point.

\paragraph{Why Joint Training?} 
To explore how our TSJT algorithm benefits the training of models, we plot the cross-entropy loss of the three strategies mentioned above.

\begin{figure*}[thb]
    \centering
    \subfigure[ConstJT-shared v.s. TSJT-shared]{
		\begin{minipage}[b]{0.46\linewidth}
		  \includegraphics[width=\textwidth]{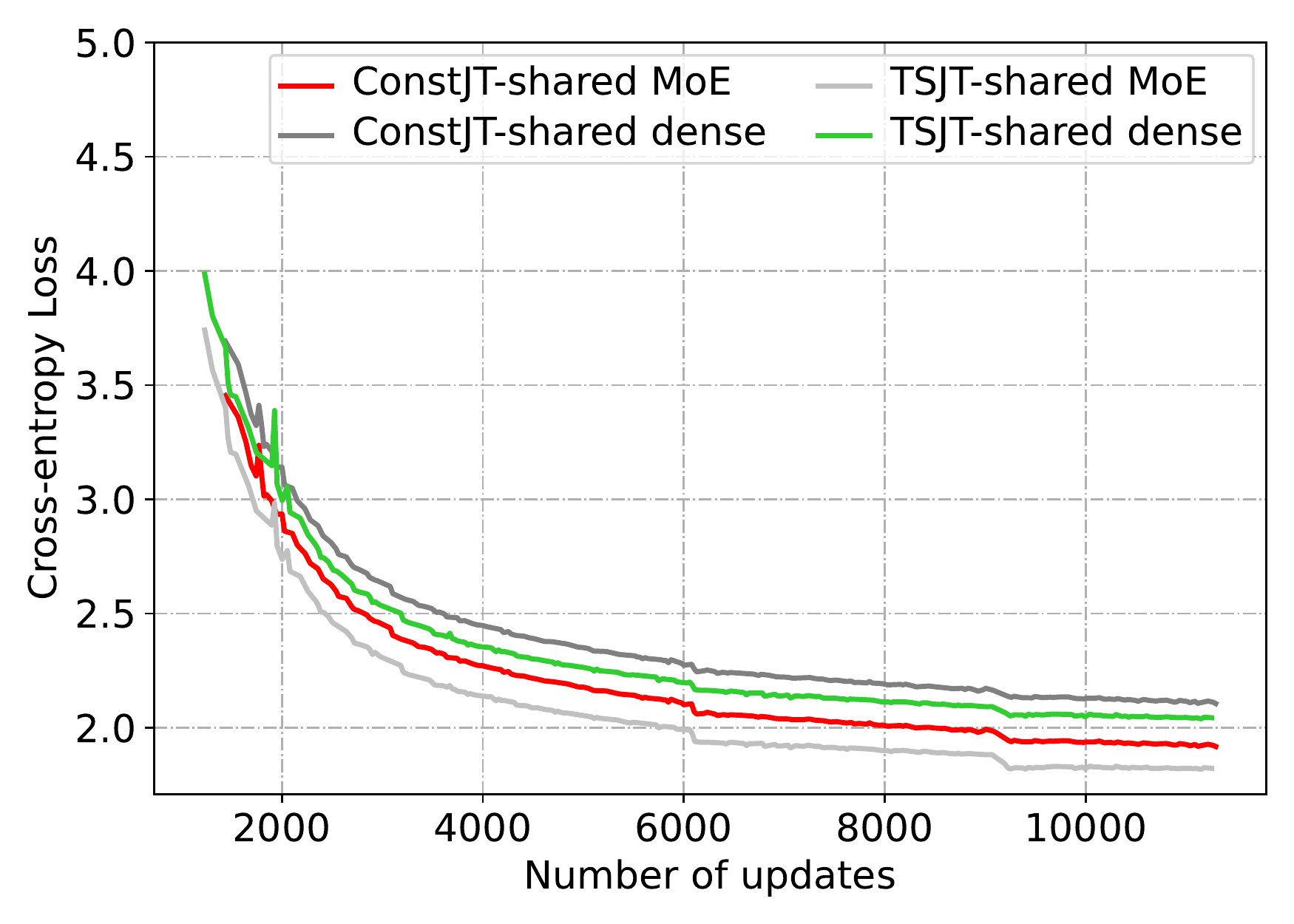}
		\end{minipage}
    }
    \subfigure[Single v.s. TSJT-shared]{
        \begin{minipage}[b]{0.46\linewidth}
          \includegraphics[width=\textwidth]{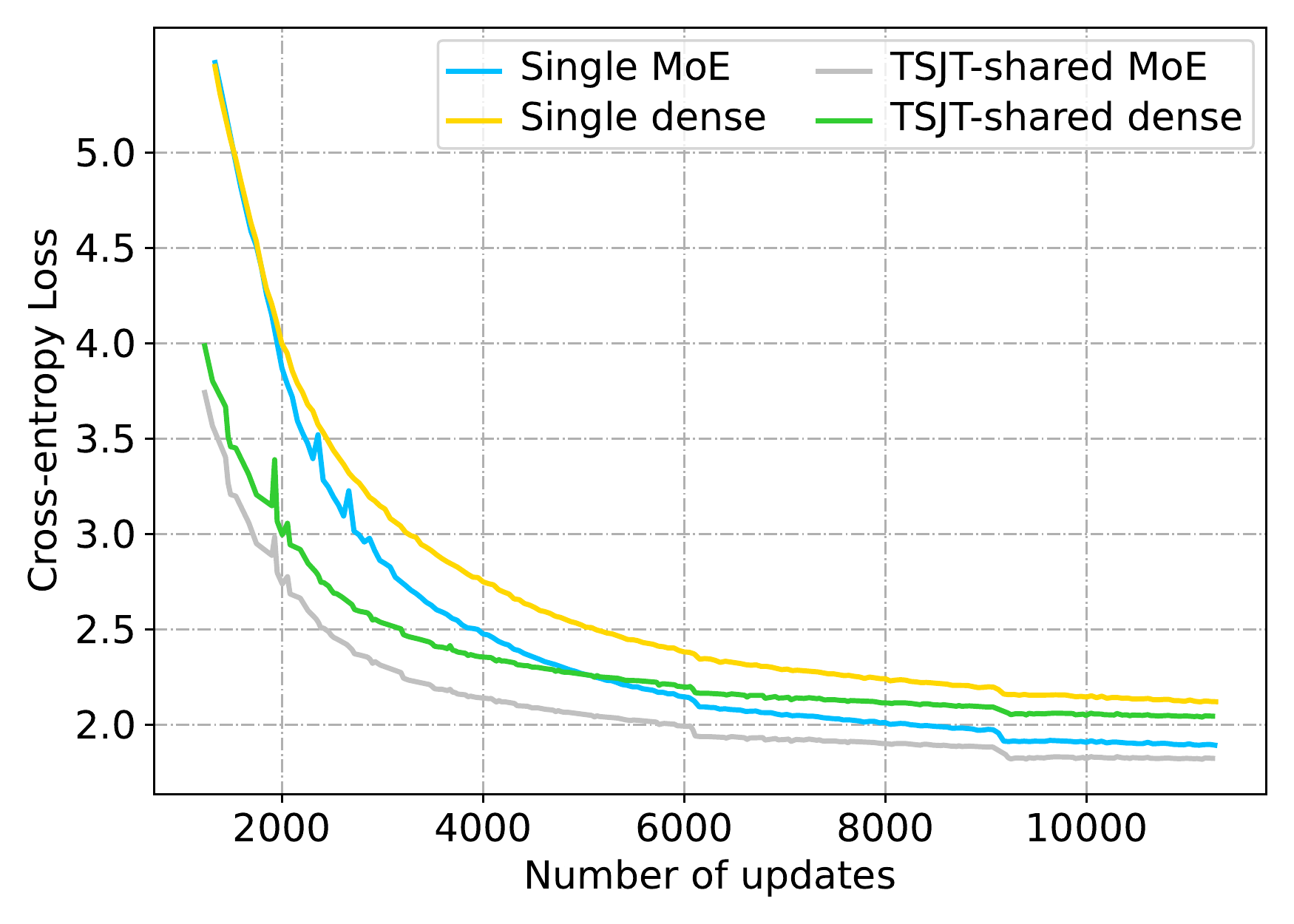}
        \end{minipage}
    }
    \subfigure[ConstJT-indep v.s. TSJT-indep]{
		\begin{minipage}[b]{0.46\linewidth}
		  \includegraphics[width=\textwidth]{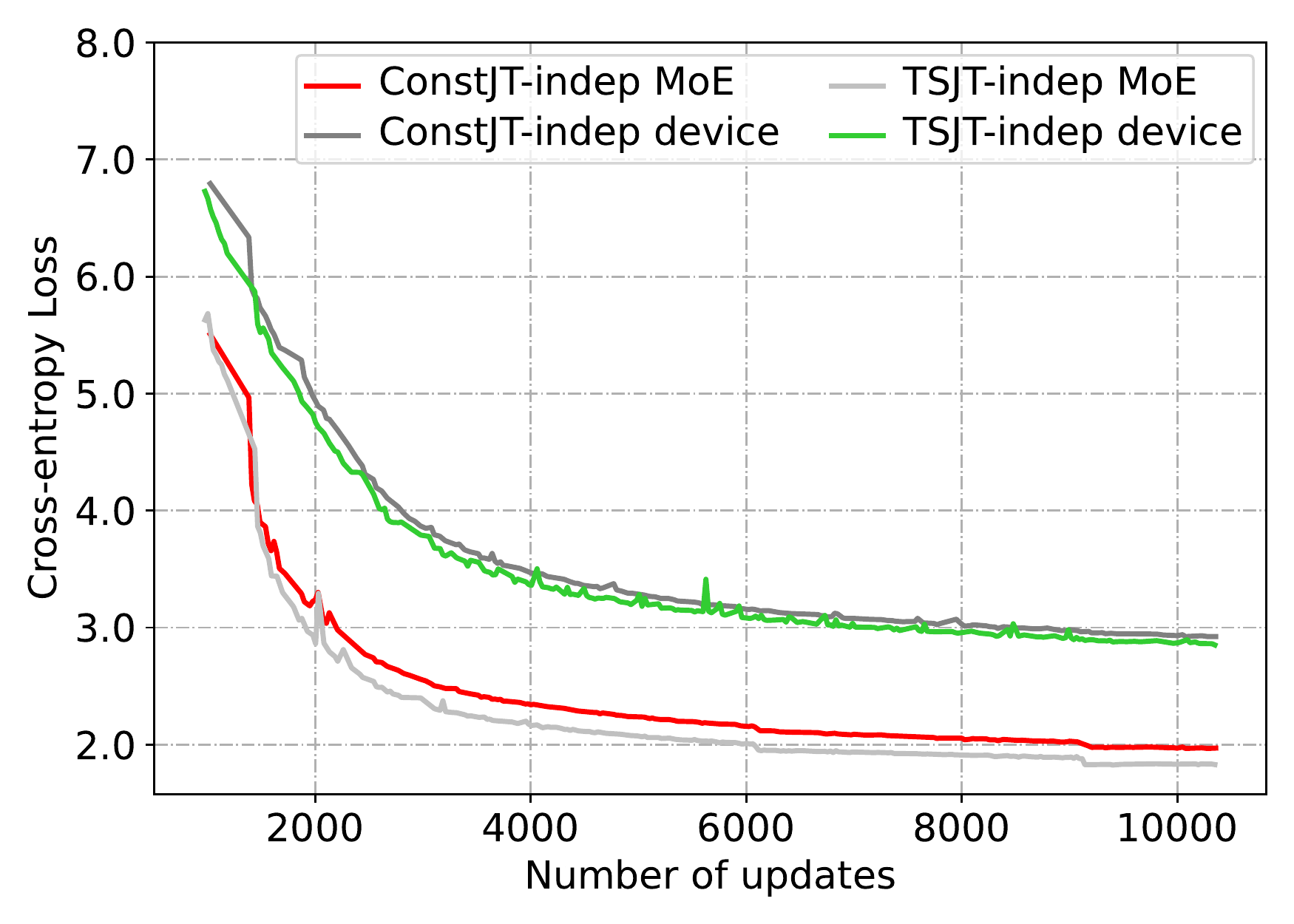}
		\end{minipage}
    }
    \subfigure[Single v.s. TSJT-indep]{
        \begin{minipage}[b]{0.46\linewidth}
          \includegraphics[width=\textwidth]{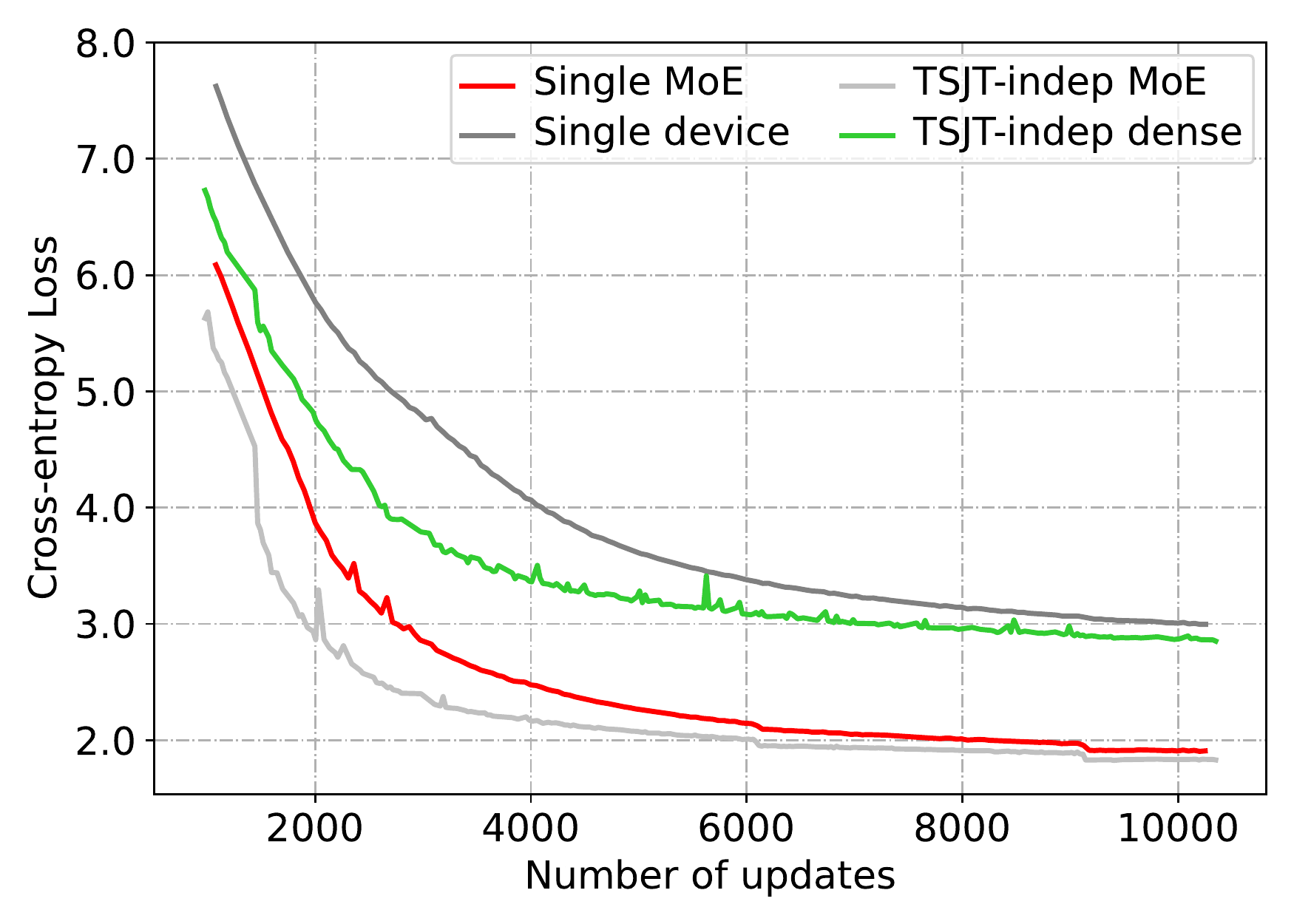}
        \end{minipage}
    }
    \caption{Optimization trajectory of models training with different algorithms.
    "Single" denotes the vanilla training, "ConstJT" denotes the constant joint-training algorithm, and "TSJT" denotes the two-stage joint-training algorithm.
    Shared and indep denotes the model architecture.}
    \label{fig:loss}
\end{figure*}


The visualizations are shown in Figure~\ref{fig:loss}.
We can observe that the TSJT approach exerts a significant impact on the optimization trajectory, particularly at the outset of the training process. 
In comparison to the single training approach, the loss of models trained using our two-stage joint-training approach decreases more rapidly and reaches a lower point in a shorter period of time.
Furthermore, the TSJT approach maintains a stable lower loss status as the training progresses. 
Despite that the constant joint-training approach also produces some benefits initially, the KL constraint ultimately has a negative effect and impedes the models from discovering optimal solutions.
This indicates that our two-stage joint-training approach can lead the models towards an efficient optimization direction by correcting each other through the KL constraint in the first stage, and subsequently, in the second stage, allows the models to individually find their optimal point while maintaining the advantage gained before.

\paragraph{Why Two Stage?} 
As previously stated, it's impractical to impose KL constraints throughout the entire training process. 
Therefore, we delved deeper into the KL loss between the MoE and the dense or device model in ConstJT and TSJT frameworks to monitor its progression over time. 
Subsequently, we plot the curve of KL loss between the MoE and dense model in the above two frameworks during the entire training process.

\begin{figure}[hbt]
    \centering
    \includegraphics[width=\linewidth]{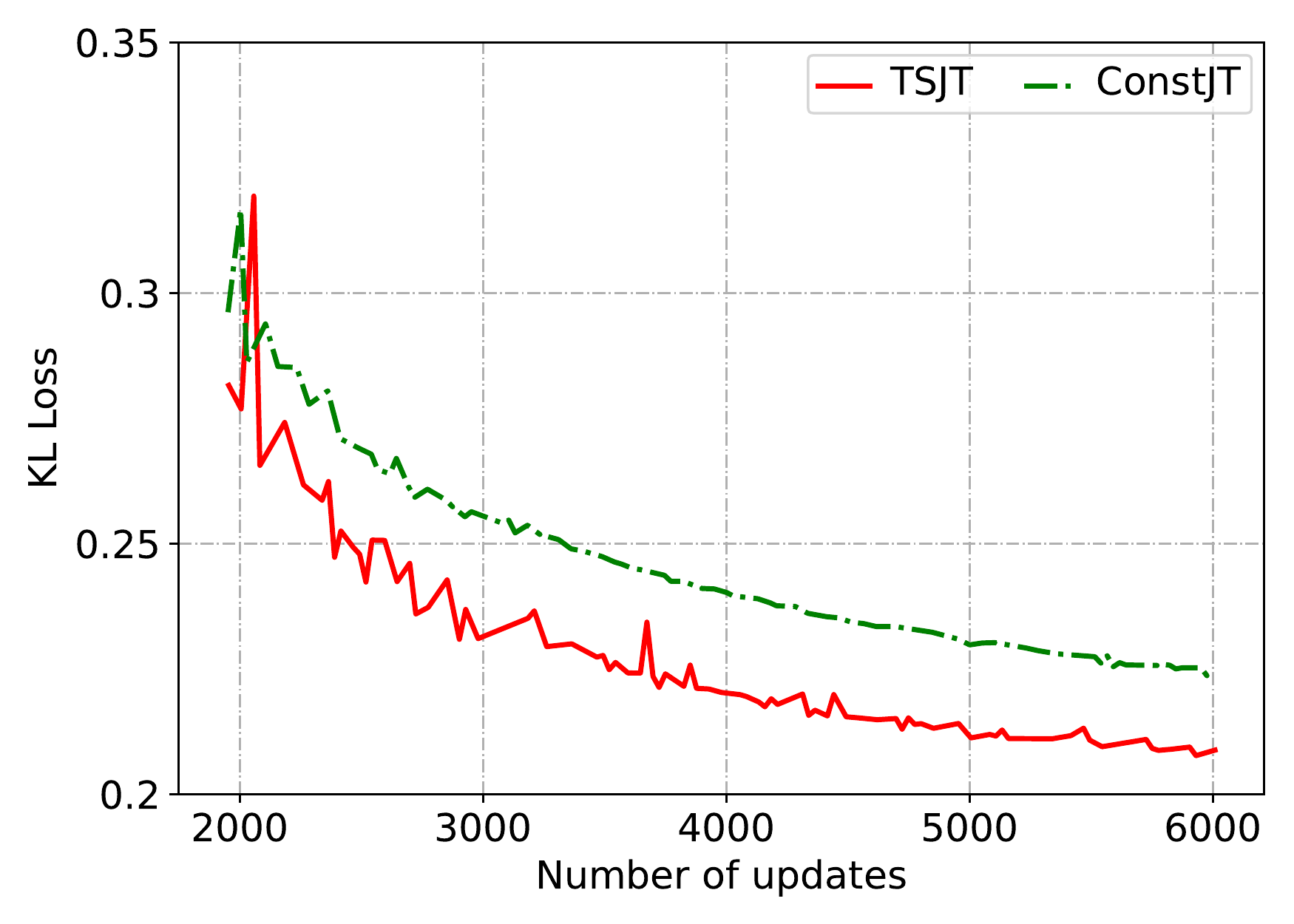}
    \caption{Kullback-Leibler (KL) loss along training process of models trained with different algorithms.}
    \label{fig:kl-loss}
\end{figure}

Results are shown in Figure~\ref{fig:kl-loss}.
We can observe that, as the number of updates increases, the KL loss of the TSJT algorithm progressively decreases, surpassing that of the ConsJT approach.
Our TSJT approach is validated by this result, as it reinforces the notion that constraints shouldn't be enforced throughout the entire training process. 
Additionally, models with varying capacities exhibit improved consistency and discover optimal solutions under TSJT, whereas models trained using ConstJT experience counterproductive outcomes.
The visualization further corroborates the findings in Table~\ref{tab:x2x-3langs}, where ConstJT initially showed encouraging results but failed to maintain its momentum as the training progressed.
\section{Related Work}

\textbf{Mixture of Experts.} 
Mixture-of-Experts (MoE) models which has been proposed about thirty years ago~\cite{6797059,10.1162/neco.1994.6.2.181} got rejuvenated recently.
MoE is usually such a neural network architecture that includes several experts.
During training and inference, MoE models route input examples to specific expert(s).
Thus different experts are learned to handle specific sets of examples.
In this way, the model size is expanded exponentially with relatively low computational cost.
MoE models have been widely applied to various domains, such as computer vision~\cite{ruiz2021scaling} and speech recognition~\cite{you21_interspeech}.
In natural language processing, recent studies focus on integrating MoE into Transformers model~\cite{NIPS2017_3f5ee243}.
GShard~\cite{lepikhin2021gshard} and Switch Transformers~\cite{JMLR:v23:21-0998} scale the original Transformers by replacing the feed-forward layers with experts layers.
MoE models have achieved state-of-art performances on various natural language processing tasks, especially neural machine translations~\cite{dai-etal-2022-stablemoe,chi2022on}.

However, the extremely high requirement for device and computation resources prevents MoE models from being widely applied to production.
Several studies~\cite{pmlr-v162-rajbhandari22a,xue2022student} explore reducing the time and computation cost of MoE models through tensor parallelism, knowledge integration and so on.

\textbf{Model Compression.}
As the scale of deep neural networks grows substantially, model compression has raised great attention in recent years.
Numerous studies aim to address the major challenge of deploying large-scale models in practical scenarios.
The most popular techniques of model compression include parameter sharing \cite{conneau-etal-2020-unsupervised}, pruning \citet{Fan2020Reducing}, quantization \cite{9463531} and knowledge distillation \cite{hinton2015distilling}.
Knowledge Distillation (KD) is one of the most common methods, which transfer knowledge of a large teacher model to a small student model.
To ensure effective knowledge transfer, KD typically involves a loss function that minimizes the distance between the output of the teacher and student models.
Depends on the optimization target, the knowledge distillation method can be roughly categorized into the logit-based KD and feature-based KD \cite{xu2022survey}.
Logit-based KD aims to align the logits of the teacher and student model.
For example, DistilBERT \cite{sanh2020distilbert} distills BERT in the pre-training stage using a carefully designed loss function that comprises the initial MLM loss, cosine similarity loss, and KL divergence.
MixKD \cite{liang2021mixkd} leverages mixup which encourages the student model to mimic the teacher's behavior on the linear interpolation of example pairs as well.
Feature-based KD \cite{jiao-etal-2020-tinybert,liu-etal-2022-multi-granularity} is similar with the logit-based KD, but it further capitalizes more knowledge from the intermedia features from the teacher models.
And all of the existing methods are not designed in parallel, \textit{i.e.}, the student model can only be obtained after the teacher model is trained.
\section{Conclusions and Future Work}
In this work, we propose a novel one-stop training schema of multiple capacity models.
Concretely, we design two composite model architectures to provide various-capacity models with flexible depth and width.
To train different-capacity submodel exhaustively at the same time, we then propose a two-stage joint training algorithm called TSJT.
It adjusts the consistency constraint at different stages.
Experimental results indicate the effectiveness of our schema, and further analysis reveals the inner working of our TSJT.

\section*{Limitations}
Although our method demonstrates success on WMT10 benchmark, it is not without limitations.
First, due to the limitation of computation resources, we only test our method on encoder-decoder based models and machine translation tasks.
In the future, we plan to expand our framework to encompass additional model backbones, such as encoder-only and decoder-only architectures, as well as other tasks like understanding and language modeling.
Moreover, the models we uesed are all trained from scratch, but our framework could also be applied to pre-trained models.
More exploration on this direction will be better.
Second, there are some vital hyper-parameters in our framework, \textit{e.g.} the separate threshold $t_\text{sep}$ in TSJT algorithm and the scaling coefficient $\alpha$ in composite training objective Eq.\ref{eq:2}.
We adopt grid search to select the best parameters, which requires considerable GPU resources.
An automatic method would be more desirable. 

\bibliography{anthology}
\bibliographystyle{acl_natbib}

\end{document}